\documentclass[sigconf, review=false]{acmart}

\usepackage{booktabs} 

\setcopyright{rightsretained}
\usepackage{multirow}
\usepackage{makecell}
\usepackage{tabularx}

\copyrightyear{2024}
\acmYear{2024}
\setcopyright{othergov}
\acmConference[ICVGIP 2024]{Indian Conference on Computer Vision Graphics and Image Processing}{December 13--15, 2024}{Bengaluru, India}
\acmBooktitle{Indian Conference on Computer Vision Graphics and Image Processing (ICVGIP 2024), December 13--15, 2024, Bengaluru, India}
\acmPrice{}
\acmDOI{}
\acmISBN{}

\begin{document}
\title{In the Era of Prompt Learning with Vision-Language Models}
\titlenote{Produces the permission block, and
  copyright information}


\author{Ankit Jha}\email{ankitjha16@gmail.com}
 \affiliation{
   \institution{Department of Computer Science and Engineering, LNMIIT}
   \city{Jaipur}
  \state{Rajasthan}
  \country{India}
 }

\renewcommand{\shortauthors}{}

\begin{abstract}
Large-scale foundation models like CLIP have shown strong zero-shot generalization but struggle with domain shifts, limiting their adaptability. In our work, we introduce \textsc{StyLIP}, a novel domain-agnostic prompt learning strategy for Domain Generalization (DG). StyLIP disentangles visual style and content in CLIP's vision encoder by using style projectors to learn domain-specific prompt tokens and combining them with content features. Trained contrastively, this approach enables seamless adaptation across domains, outperforming state-of-the-art methods on multiple DG benchmarks. Additionally, we propose AD-CLIP for unsupervised domain adaptation (DA), leveraging CLIP’s frozen vision backbone to learn domain-invariant prompts through image style and content features. By aligning domains in embedding space with entropy minimization, AD-CLIP effectively handles domain shifts, even when only target domain samples are available. Lastly, we outline future work on class discovery using prompt learning for semantic segmentation in remote sensing, focusing on identifying novel or rare classes in unstructured environments. This paves the way for more adaptive and generalizable models in complex, real-world scenarios.

\end{abstract}

%
%

\ccsdesc[500]{Computing methodologies~Classification Clustering, Prompt Learning}

\keywords{Prompt learning, Domain generalization, Domain adaptation}

    
\maketitle

\section{Vision-Language Models}
Vision-Language Models have revolutionized the fields of computer vision and natural language processing by leveraging shared embeddings across these modalities. Models such as CLIP \cite{clip} align images with corresponding text descriptions in a common space, allowing for flexible downstream tasks such as zero-shot learning, image classification, and retrieval. The success of VLMs lies in their ability to generalize across diverse tasks and domains without fine-tuning for each specific case.

The integration of prompt learning into VLMs is a natural evolution that enables fine-tuned adaptation to specific domains and tasks, offering opportunities for efficient transfer learning. Instead of retraining the model, task-specific prompts guide the model's focus, ensuring alignment with the desired domain or objective.

\section{Current Research}
Domain generalization and adaptation are crucial challenges in computer vision, particularly for applications that must operate across varied and unseen visual distributions. Traditional approaches often struggle when tested on out-of-distribution data, leading to degraded performance. Prompt learning addresses this by creating adaptable instructions or hints that guide the VLM to focus on domain-invariant features. Research has shown that prompt-based models, such as those inspired by CoOp \cite{coop}, CoCoOp \cite{cocoop}, etc., can be fine-tuned to specific tasks while retaining the generality of the underlying VLM. For instance, learning textual prompts allows the model to perform well on image classification tasks across domains by leveraging semantic relationships between classes and domains, thereby increasing robustness to domain shifts. Our works in this area involves learning optimal prompts that generalize across multiple domains. By leveraging pre-trained VLMs, we aim to minimize the gap between training and testing distributions, thus improving performance without extensive retraining. This has shown promising results in remote sensing applications, where visual characteristics can vary widely depending on geographic, temporal, or sensor conditions.
\subsection{Domain Generalization}
Recent advances in vision-language models like CLIP has transformed computer vision by aligning visual and textual data in a shared space, enabling strong zero-shot generalization. However, designing effective prompts remains challenging, and methods like CoOp \cite{coop} struggle with domain shift, where models fail to generalize across varying data distributions. Domain Generalization (DG) \cite{stylip,odgclip} addresses this by learning domain-agnostic features, but traditional DG methods rely solely on image-based encoders. While prompt engineering with CLIP shows promise, current approaches lack robustness with large domain variations.
\begin{table}[!ht]
    \scriptsize{
    \begin{center}
    
    \caption{Comparison of our proposed \textsc{StyLIP} with the state-of-the-art methods on PACS, VLCS, Digits-DG, Office-Home, and DomainNet datasets for multi-source DG in terms of mean leave-one-out performance.}
    \vspace{-0.32cm}
    \scalebox{1.2}{
    \begin{tabular}{lccccc}
    \toprule
    \textbf{Method} & \textbf{PACS}& \textbf{VLCS}& \textbf{Off.Home}&\textbf{Dig.DG}&\textbf{Dom.N.}\\ 
    \midrule
        MaPLe \cite{maple} & 97.56 & 85.12 & 83.35 & - & 60.43 \\
        \textsc{StyLIP}&\textbf{98.05}&\textbf{86.94}  &\textbf{84.63} &\textbf{81.38}&\textbf{62.02}\\ 
         \bottomrule
    \end{tabular}}
    \label{tab:1}
    \end{center}}
    \vspace{-0.3cm}
\end{table}
We propose \textsc{StyLIP} \cite{stylip}, a prompt-tuning strategy for CLIP that conditions prompts on domain and content information from visual features. By leveraging instance-wise statistics, \textsc{style projectors} learn domain-specific tokens to enhance prompts. Additionally, object-level variations improve cross-domain generalization. Unlike prior methods \cite{coop, cocoop}, \textsc{StyLIP} fuses multi-layer visual features with prompt embeddings, achieving superior results in DG scenarios, as shown in Table \ref{tab:1} on five benchmarks.

Following these, we introduce APPLeNet \cite{applenet}, a vision-language model for classifying complex remote sensing images using attention-guided prompt learning under a few-shot paradigm, with results compared on four remote sensing datasets, shown in Table \ref{tab:2}.
\begin{table}[ht!]
\centering
\scriptsize{
    \centering
    \caption{\small{Comparison of APPLeNet with state-of-the-art methods for base-to-new (B2N) class generalization task. Here, H represents the harmonic mean between the base and new class accuracies.}}
    \vspace{-0.3cm}
    \scalebox{1.07}{
    \begin{tabular}{lccc} 
    \toprule
	
&\multicolumn{3}{c}{\textbf{Avg. of all}}\\
      \cmidrule(lr){2-4}
     
 \textbf{Method} &\multicolumn{1}{c}{\textbf{Base}}&\multicolumn{1}{c}{\textbf{New}}&\multicolumn{1}{c}{\textbf{H}}\\
    
    \midrule

  CoCoOp \cite{cocoop}& 
  87.92 & 57.99 & 69.88 \\

   APPLeNet &
   \textbf{89.98} & \textbf{60.79} & \textbf{72.56} \\ \bottomrule
    \end{tabular}}\label{tab:2}}
    \vspace{-0.5cm}
\end{table}
\subsection{Domain Adaptation}
Domain adaptation (DA) addresses the challenge of transferring knowledge from a labeled source domain to an unlabeled target domain, where data distributions differ. The goal is to minimize domain shift and improve performance on the target domain. Unlike domain generalization (DG), which aims to generalize to unseen domains without access to target data, DA leverages target domain information for adaptation. Recently, prompt learning has been used to tackle DA in vision-language models like CLIP, offering a more efficient alternative to full model fine-tuning. By modifying input text prompts, models can better align visual and textual features across domains, reducing domain shift. While some DA approaches use prompt learning, most rely on fixed or heuristic token designs, limiting their flexibility in handling varying domain distributions.

We propose \textsc{AD-CLIP}, a novel framework for unsupervised DA using prompt learning in foundation models. Our method, based on CLIP, learns domain-invariant and class-generic prompt tokens by extracting multi-scale style and content features from CLIP's vision encoder. Specifically, \textsc{AD-CLIP} learns three types of tokens—domain, image, and class tokens—per image, allowing it to handle diverse domain variations. To further enhance adaptation, we introduce a combination of distribution divergence and entropy minimization losses for effective source-target domain alignment. Experiments on three benchmark DA datasets show that \textsc{AD-CLIP} outperforms state-of-the-art methods, highlighting the power of prompt learning in domain adaptation, shown in Table \ref{tab:3}.

We are exploring how domain adaptation (DA) and domain generalization (DG) can enhance various computer vision tasks. For example, in visual-question answering (VQA), we aim to improve models' ability to interpret images and text across different environments, allowing accurate responses without extensive retraining. In multimodal learning-based speech recognition, we investigate how DA and DG can align audio-visual features despite variations in environments and accents, enhancing robustness in real-world applications. Our research focuses on developing models that adapt to new environments with minimal supervision, reducing the need for training from scratch and limiting the number of parameters. This is especially valuable for tasks with small sample sizes or scarce labeled data. Ultimately, we aim to harness DA and DG to enable foundation models to generalize effectively across a range of computer vision tasks while maintaining learning efficiency.
\begin{table}[ht!]

    \centering
    \caption{Comparison of \textsc{AD-CLIP} with state-of-the-art methods for UDA task on Office-Home dataset.}
    \vspace{-0.5cm}
    \centering
    \scalebox{0.9}{
    \begin{tabular}{lc}
    \toprule
        
       {\textbf{Method}} & {\textbf{Avg Accuracy}} \\ 
        \midrule
        DAPL \cite{dapl}
        & $88.7$ \\



         
        \textsc{AD-CLIP} &  $\textbf{90.5}$ \\\hline
        
        
    \end{tabular}
    \label{tab:3}
    }
    \vspace{-0.5cm}
\end{table}
\section{Future Directions}
While current research on prompt learning in VLMs focuses predominantly on image classification, we propose extending this approach to class discovery in semantic segmentation tasks. In remote sensing, identifying new or rare classes without prior knowledge of their existence is a significant challenge. Class discovery, particularly in unstructured environments, can benefit from prompt learning by leveraging learned semantic relationships from existing classes to infer new ones. Prompt learning could be adapted to segmentation by providing context-specific prompts that help the model focus on underrepresented or previously unseen classes. This would be particularly useful in remote sensing, where class distributions are often long-tailed, and rare or emerging classes are of high importance for applications such as environmental monitoring, disaster management, or urban planning. Incorporating prompt learning in segmentation models could enable them to not only classify known regions but also discover and delineate novel semantic classes, improving their utility in real-world tasks that require flexibility and adaptability. Future work will focus on designing prompt structures that can guide segmentation models toward identifying such new classes, leveraging both visual and textual cues.

\bibliographystyle{ACM-Reference-Format}
\bibliography{cite}





\end{document}